\documentclass[10pt,twocolumn,letterpaper]{article}

\usepackage{cvpr}
\usepackage{amsmath}
\usepackage{array}
\usepackage{amssymb}
\usepackage{amsfonts}
\usepackage{amsthm}
\usepackage{mathrsfs}
\usepackage{graphicx,subfigure}
\usepackage{cite}
\usepackage{epstopdf}
\usepackage{caption}
\usepackage{algorithm,algorithmicx}
\usepackage{booktabs}
\usepackage{multirow}

\graphicspath{{/}}

\usepackage[pagebackref=true,breaklinks=true,letterpaper=true,colorlinks,bookmarks=false]{hyperref}

\cvprfinalcopy
\def\cvprPaperID{****}

\ifcvprfinal\fi

\begin{document}

\title{HPILN: A feature learning framework for cross-modality person re-identification}

\author{Jian-Wu Lin$^{1}$, Hao Li$^{2}$
\thanks{$^*$Corresponding author. E-mail: jianwu.lin@foxmail.com }
\thanks{The authors are with College of Information Engineering, Zhejiang University of Technology, Hangzhou 310023, China.}
}

\maketitle

\begin{abstract}
Most video surveillance systems use both RGB and infrared cameras, making it a vital technique to re-identify a person cross the RGB and infrared modalities. This task can be challenging due to both the cross-modality variations caused by heterogeneous images in RGB and infrared, and the intra-modality variations caused by the heterogeneous human poses, camera views, light brightness, etc. To meet these challenges a novel feature learning framework, HPILN, is proposed. In the framework existing single-modality re-identification models are modified to fit for the cross-modality scenario, following which specifically designed hard pentaplet loss and identity loss are used to improve the performance of the modified cross-modality re-identification models. Based on the benchmark of the SYSU-MM01 dataset, extensive experiments have been conducted, which show that the proposed method outperforms all existing methods in terms of Cumulative Match Characteristic curve (CMC) and Mean Average Precision (MAP).
\end{abstract}


\section{Introduction}

Person re-identification (Re-ID) is the technique of identifying an individual from a surveillance camera who has previously shown up from other non-overlapping cameras \cite{zheng2016person}, which has recently become a research hotspot in the field of computer vision due to its practical importance. Typical Re-ID uses only RGB cameras, i.e., identifying an individual from RGB cameras based on previously recorded RGB camera videos/images, and hence the name RGB-RGB Re-ID\cite{yu2017devil,sun2018beyond,chang2018multi,dai2018batch,wang2018learning}. However, in many cases both RGB and infrared cameras are used, and consequently it is necessary to develop Re-ID methods capable of cross RGB and infrared modalities, that is, either identifying an individual from RGB cameras based on previously recorded infrared camera videos/images, or identifying an individual from infrared cameras based on previously recorded RGB cameral videos/images, both being referred to as RGB-IR Re-ID \cite{wu2017rgb, ye2018visible, dai2018cross,kang2019person,Wang_2019_CVPR}.

\begin{figure}[!t]
  \centering{\includegraphics[width=0.9\linewidth]{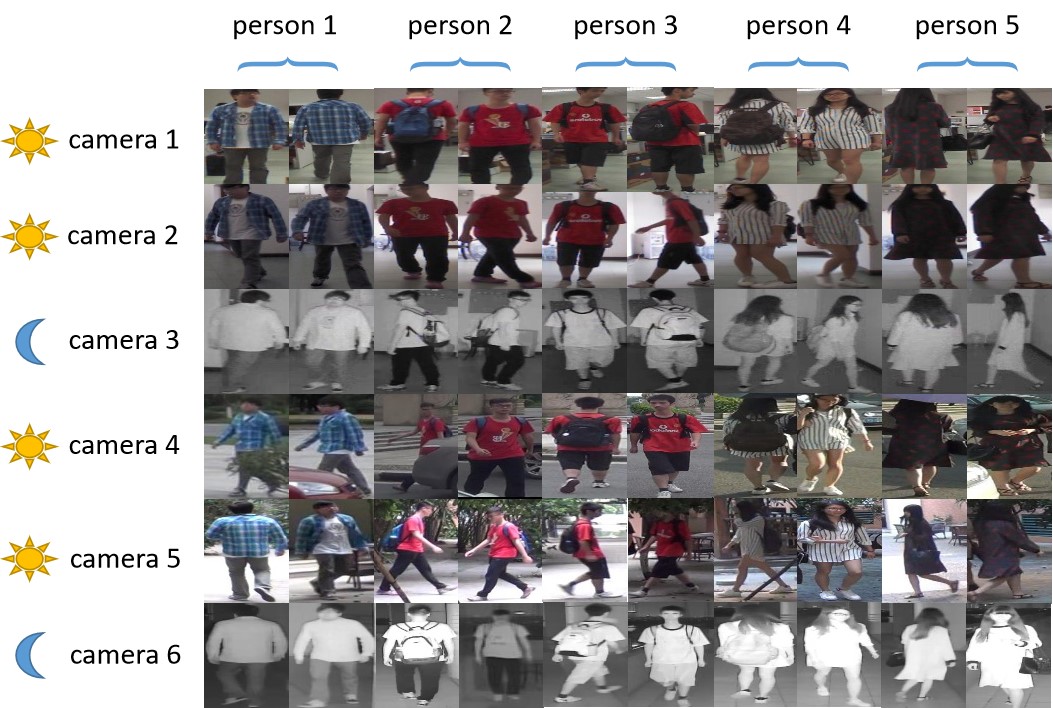}}
  \caption{RGB and infrared images in SYSU-MM01 dataset. The $1^{st}, 2^{st}, 4^{st}, 5^{st}$ and $3^{st}, 6^{st}$ rows are captured by RGB and infrared cameras, respectively
  \label{fig:SYSU}}
\end{figure}

RGB-IR Re-ID has not been well studied to date, with few literature being reported. To name just a few, in \cite{wu2017rgb}, a deep zero-padding network is proposed to automatically learn the common features of the two modalities. In \cite{ye2018visible} a dual-path network with top-ranking loss is proposed which  considers both the cross-modality and intra-modality variations. In \cite{dai2018cross} a cmGAN approach with cross-modality triplet loss is proposed to learn the discriminative feature. In \cite{kang2019person} a single image input method is proposed to simplify the convolutional neural network structure. In \cite{Wang_2019_CVPR} a dual-level discrepancy reduction learning (D$^{2}$RL) scheme is  proposed to decompose the mixed modality and appearance discrepancies. A dedicated dataset for RGB-IR Re-ID called SYSU-MM01 has  been collected \cite{wu2017rgb}, as shown in Fig. \ref{fig:SYSU}.

\begin{figure}[!t]
\centering{
\subfigure[cross-modality variations]{
\begin{minipage}[t]{0.5\linewidth}
\centering
\includegraphics[width=0.9\linewidth]{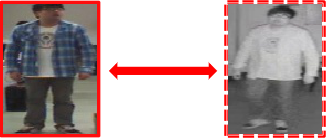}
\label{subfig:outward-b}
\end{minipage}
}
\subfigure[intra-modality variations]{
\begin{minipage}[t]{0.5\linewidth}
\centering
\includegraphics[width=0.9\linewidth]{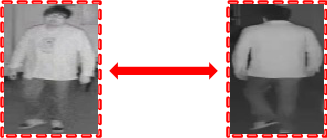}
\label{subfig:outward-d}
\end{minipage}%
}}%
\caption{Cross-modality and intra-modality variations in RGB-IR Re-ID. The solid and the dotted lines are for the RGB and infrared domains, respectively. \label{fig:diff-RGBIR}}
\end{figure}

RGB-IR Re-ID is challenging mainly due to the great cross-modality and intra-modality variations as illustrated in Fig.\ref{fig:diff-RGBIR}. By "cross-modality variation" we mean that RGB and infrared images are essentially heterogeneous as the former consists of  three channels of color information while the latter only one. By "intra-modality variation" we mean that the image quality including the camera view, resolution, light brightness, human body pose, etc. can still be significantly different even within the same RGB or infrared modality, as long as multiple heterogeneous cameras and different monitoring scenarios are involved.

To meet the above challenges,  a novel feature learning framework based on
hard pentaplet and identity loss network (HPILN) is proposed in this work.
Specifically, we select existing RGB-RGB Re-ID models as the feature
extraction module in our framework
\cite{yu2017devil,sun2018beyond,chang2018multi,dai2018batch,wang2018learning},
and then design the hard pentaplet loss to compensate for the deficiencies
of the RGB-RGB Re-ID model in the cross-modality Re-ID task. The hard
pentaplet loss considers the following two aspects: 1) a pentaplet loss,
consisting of the global and cross-modality triplet loss where the former
can simultaneously handle cross-modality and intra-modality variations, and
the latter can increase the ability to handle cross-modality variations. 2)
an improved hard mining sampling method by selecting the hardest global
triplet and the hardest cross-modality triplet to form the hardest pentaplet
pair and to contribute to the convergence of the convolutional neural
networks.

The main contributions of this paper can be summarized as follows.
\begin{itemize}
\item An end-to-end feature learning framework is proposed yielding the state-of-the-art performance on the RGB-IR Re-ID dataset SYSU-MM01.
\item The proposed RGB-RGB Re-ID model migration to RGB-IR Re-ID task provides a superior feature extraction method for future improvements.
\item A novel loss function called hard pentaplet loss is proposed which
    is capable of simultaneously handling the cross-modality  and
    intra-modality variations in RGB-IR Re-ID.
\end{itemize}

The remainder of the paper is organized as follows. Section \ref{sec:RelatedWork} provides some preliminaries on Re-ID. The proposed method is detailed in Section \ref{sec:ProposedMethod}, which is then verified experimentally in Section \ref{sec:Experimental}. Section \ref{sec:Conclusion} concludes the paper.

\section{Related Work}\label{sec:RelatedWork}

In this section we discuss related works on  single-modality and multi-modality Re-ID.

\subsection{Single-modality person re-identification}

In the single-modality person re-identification study, most attentions have been paid to RGB-RGB Re-ID.

For RGB-RGB Re-ID, hand-designed descriptors are often used to extract pedestrian features such as color and texture information. In \cite{farenzena2010person}, pedestrian body is segmented from the background, and then the weighted color histogram and the maximally stable color regions are calculated for the pedestrian body part. In recent years, the mainstream of Re-ID is to design the loss function and convolutional neural networks based on deep learning methods. The design of the loss function may depend on either metric learning or representation learning. Metric learning aims to learn the similarity of two pedestrian images through a deep CNN network, where the similarity is usually represented by the Euclidean distance. Frequently used metric learning methods include contrastive loss \cite{varior2016gated}, triplet loss \cite{schroff2015facenet}, hard triplet loss \cite{hermans2017defense} and quadruplet loss \cite{chen2017beyond}. Representation learning uses identity tags to automatically extract pedestrian representation  features, including identity loss \cite{xiao2016learning} and verification loss \cite{chen2018deep}. In addition, three types of special networks have been designed for Re-ID, i.e., either global-based, or part-based, or attention-based. Global-based networks aggregate global-level features into a global vector \cite{yu2017devil, chang2018multi}. Part-based networks divide the pedestrian image into different parts, and the local feature vectors of different parts is merged into a vector \cite{wang2018learning, sun2018beyond, dai2018batch}. Attention-based networks focus on automatically finding local salient regions for computing deep features \cite{qian2017multi, li2018harmonious}. These existing single-modality re-identification models have rarely been applied to RGB-IR Re-ID to date and efforts need to be taken for such a migration.

\subsection{Multi-modality person re-identification}

Existing multi-modal fusion person re-identification focus on RGB-D modules \cite{wu2017robust, barbosa2012re}, visible-thermal modules\cite{ye2018visible, kniaz2018thermalgan} and RGB-IR modules \cite{wu2017rgb}. RGB-D Re-ID combines human RGB image and depth information, and depth information is used to provide more invariant body shape and skeleton information to reduce the impact of changed clothes or extreme illumination on re-identification. RGB-IR and visible-thermal (VT) Re-ID is based on the principle of infrared imaging, enabling re-identification to take place at night. The difference is that the RGB-IR Re-ID transmits and collects infrared light through the infrared camera to obtain infrared images, while the VT Re-ID capturing the heat emitted by the human body to obtain infrared images. However, depth cameras and thermal cameras are rare in surveillance systems. In contrast, infrared cameras have been widely deployed. Most surveillance cameras in the real world are visible light cameras during the day and become infrared cameras at night. Therefore, from the perspective of practical applications, RGB-IR Re-ID can be of more value.

\begin{figure}[!b]
  \centering{\includegraphics[width=0.9\linewidth]{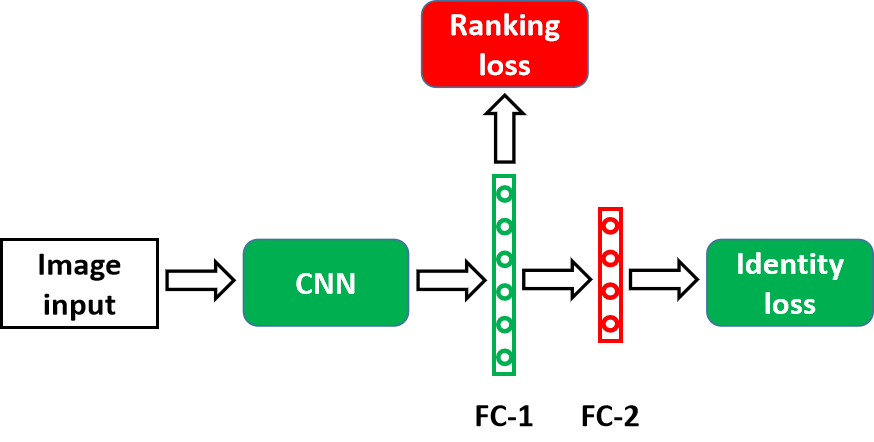}}
  \caption{A typical RGB-RGB Re-ID CNN model. The green and the red represent the unchanged and changed parts, respectively \label{fig:RGB-RGB}}
\end{figure}

\begin{figure*}[!t]
\centering{\includegraphics[width=0.9\linewidth]{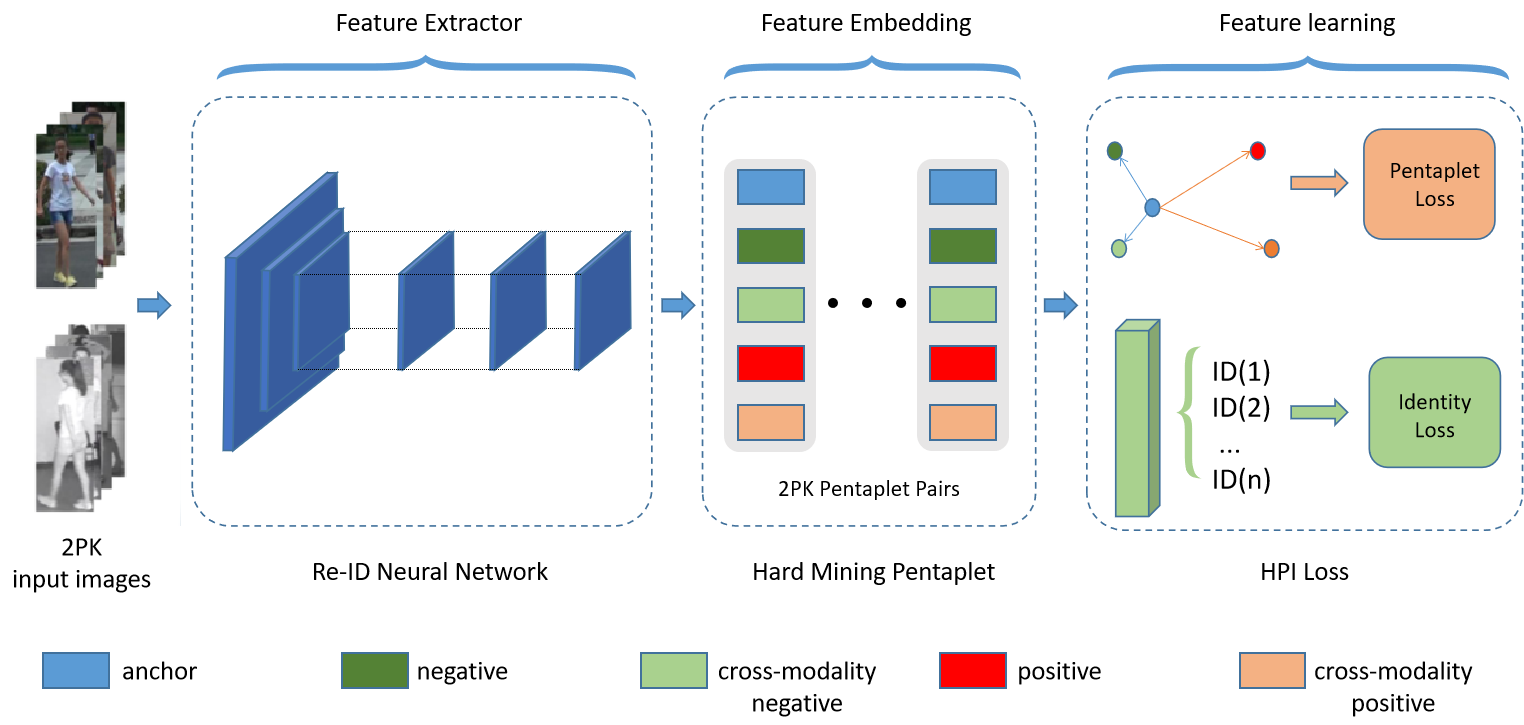}}
\caption{The proposed feature learning framework based hard pentaplet and identity (HPI) loss for RGB-IR Re-ID. The framework consists of  three main components: 1) The Re-ID neural network which extracts the common features of RGB and infrared images; 2) The hard mining sampling method which obtains the hardest pentaplet pair sets; 3) The HPI loss for feature learning which consists of pentaplet loss and identity loss. $2PK$ is the training batch size. In each training batch, $P$ individuals are randomly selected, and each person randomly selects $K$ RGB images and $K$ infrared images. The rectangles of different colors below the image represent the different elements in the pentaplet pair. \label{fig:framework}}
\end{figure*}

\section{The Proposed Method}\label{sec:ProposedMethod}

This paper addresses RGB-IR Re-ID by a feature learning framework based on hard pentaplet loss and identity loss as shown in Fig.\ref{fig:framework}. The framework consists of three parts: 1) Re-ID neural network for feature extraction; 2) the hard mining sampling method to find hardest pentaplet pair sets  after getting feature embedding; 3) HPI loss for feature learning. Specifically, the Re-ID neural network is taken from  existing RGB-RGB Re-ID convolutional neural network which can also extract the representation feature of infrared person images. By calculating the Euclidean distance of the feature embedding, the hard mining sampling method maximizes training and ensures model convergence. The hard pentaplet loss enables the network to handle cross-modality and intra-modality variations simultaneously, and the hard pentaplet loss and the identity loss are integrated into multiple losses to facilitate the process of feature learning.

\subsection{Re-ID neural network}\label{sec:ReIDNetwork}

We use RGB-RGB Re-ID neural network as feature extraction modules to extract common features of two  heterogeneous modalities, since these RGB-RGB Re-ID-specific models can outperform image classification models, despite the heterogeneous images  from two modalities.

In our framework, we slightly adjust the structure of the RGB-RGB Re-ID model. A typical RGB-RGB Re-ID model based convolutional neural network is shown in Fig.\ref{fig:RGB-RGB}. Most  Re-ID models have at least two fully connected layers, where the last layer (FC-2) is for identity loss, and the output of the penultimate layer (FC-1) is used as feature embedding supervised by ranking loss based on metric learning. In our method we change the dimension of the last fully connected layer (FC-2) to the number of person class in the SYSU-MM01 training set. The front convolutional neural network (CNN) part is used as feature extractor to obtain feature embedding.

\subsection{Hard pentaplet loss}
Our approach is inspired by the combination of hard triplet loss and triplet loss. We discuss in turn  the triplet loss, the hard triplet loss, and finally the proposed hard pentaplet loss.

\subsubsection{The triplet loss}
The triplet loss is widely used in image retrieval tasks such as face recognition, person re-identification, and vehicle retrieval. In the person re-identification task, for the anchor image $x^{a}$ in the candidate triplet set $\{x^{a}_{i},x^{p}_{i},x^{n}_{i}\},i\in[1,N]$, $x^{p}$ is a positive sample image of the same identity, and $x^{n}$ is a negative sample image of a different identity. Using the convolutional neural network as the feature extractor, the image $x$ is mapped into the $d$-dimensional Euclidean space. The feature embedding vector can be expressed as $f(x)\in\mathbb{R}^d$. The Euclidean distance between feature embedding measures the similarity of two images, which can be expressed as follows,
\begin{align}
     d(x_{i},x_{j})=\| f(x_{i})-f(x_{j}) \Vert_{2}
\end{align}

The triplet loss is obtained as follows,
\begin{align}\label{TripletLoss}
     L_{trp}=\sum_{i}^{N}[d(x^{a}_{i},x^{p}_{i})^{2}-d(x^{a}_{i},x^{n}_{i})^{2}+ \alpha]_{+}
\end{align}
where $[z]_{+}=max(z,0)$. For $\{x^{a}_{i},x^{p}_{i},x^{n}_{i}\}$, the $i$-th pair of triplets, $d(x^{a}_{i},x^{p}_{i})$ represents the Euclidean distance between positive samples $(x^{a}_{i},x^{p}_{i})$, and $d(x^{a}_{i},x^{n}_{i})$ represents the Euclidean distance between negative samples $(x^{a}_{i},x^{n}_{i})$. $\alpha$ is a hyperparameter that forces the positive and negative sample pairs to separate in the Euclidean space.

Under the supervision of triple loss, the CNN  can learn discriminative feature embedding in Euclidean space. It can be seen from Equation \eqref{TripletLoss} that if the positive sample becomes larger or the negative sample becomes smaller, the loss value will increase, and the adjustment of the weight and bias of the CNN will be larger during the back propagation. Intuitively, the triplet loss reduces the distance between positive samples, i.e., the intra-class distance, increases the distance between negative samples, i.e., the inter-class distance, and finally distinguishes different person in the Euclidean space.

The training goal is that for any triplet $\{x^{a}_{i},x^{p}_{i},x^{n}_{i}\}$, the positive and negative sample pairs in the Euclidean space meet the following inequality,
\begin{align}\label{TripletGoal}
     d(x^{a}_{i},x^{p}_{i})^{2} + \alpha < d(x^{a}_{i},x^{n}_{i})^{2}
\end{align}

\subsubsection{The hard triplet loss}
In order to ensure the network convergence, it is necessary to choose triplets that violate  \eqref{TripletGoal}. Let the triplet that already satisfies  \eqref{TripletGoal} be named by ``easy triplet". It is then not wise to randomly choose a triplet set since it would contain many such easy triplets and hence harm  the convergence of the model.

\begin{figure}[!t]
\centering
\subfigure[Hard triplet loss]{
\begin{minipage}[t]{0.9\linewidth}
\centering
\includegraphics[width=1\linewidth]{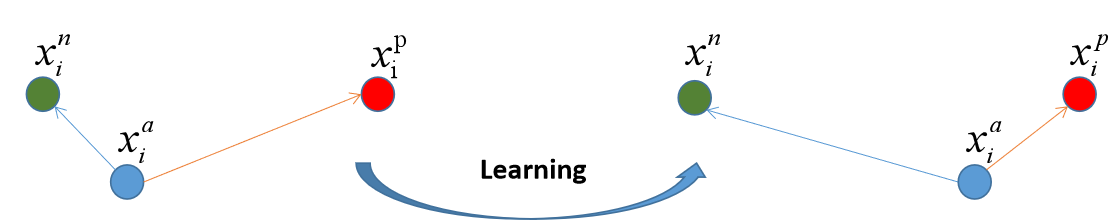}
\label{subfig:tripletloss}
\end{minipage}%
}%

\subfigure[Hard pentaplet loss]{
\begin{minipage}[t]{0.9\linewidth}
\centering
\includegraphics[width=1\linewidth]{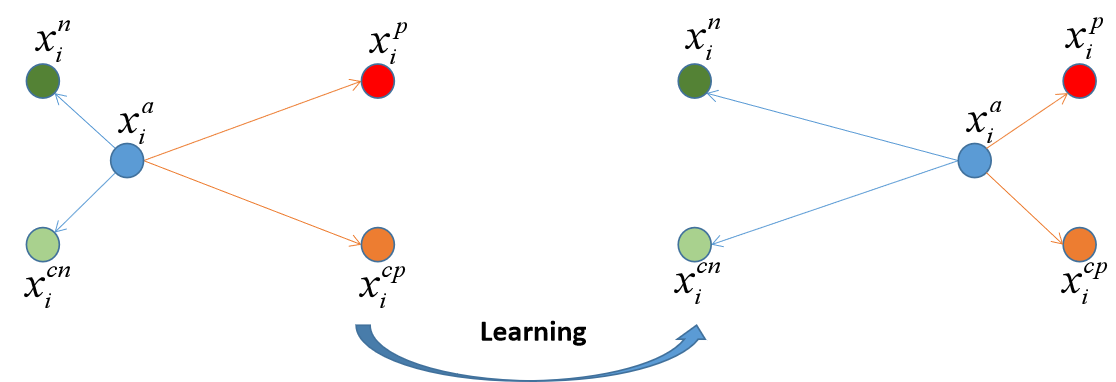}
\label{subfig:pentapletloss}
\end{minipage}%
}%
\caption{Geometry interpretation of triplet loss and pentaplet loss in
Euclidean space. (a) The triplet loss minimizes the distance between the
anchor $x_{i}^{a}$ and a positive $x_{i}^{p}$, and maximizes the distance
between the anchor $x_{i}^{a}$ and a negative $x_{i}^{n}$. (b) In addition
to the function of triplet loss, pentaplet loss can minimize the distance
between an anchor $x_{i}^{a}$ and a cross-modality positive $x_{i}^{cp}$,
and maximizes the distance between the anchor $x_{i}^{a}$ and a
cross-modality negative $x_{i}^{cn}$. \label{fig:tri-penta}}
\end{figure}

Alexander Hermans et al. designed the hard triplet loss
\cite{hermans2017defense}, which improves the training speed and accuracy in
many retrieval tasks by improving the triplet sampling method. Using this
loss, each batch randomly samples $P$-identity person, and each person
randomly samples $K$ images, thus  $PK$ images for each batch. For each
sample in the batch, select the hardest positive and negative samples to
form the hardest triplet. The hardest positive sample represents the
positive sample with the largest Euclidean distance from the anchor, and the
hardest negative sample represents the negative sample with the smallest
Euclidean distance from the anchor. The hard triplet loss can be expressed
as follows,
\begin{align}\label{HardTriplet}
    L_{htrp}=\overbrace{\sum_{i=1}^{P}\sum_{a=1}^{K}}^{all\ anchor}\
    [\alpha + \overbrace{\max_{p=1\dots K}d(x^{a}_{i},x^{p}_{i})}^{hardest\ positive}& \notag \\
    -\underbrace{\min_{\substack {j=1\dots P\\ n=1\dots K\\j\neq i }}d(x^{a}_{i},x^{n}_{j})}_{hardest\ negative}&]_{+}
\end{align}

\subsubsection{The hard pentaplet loss}
As shown in Fig.\ref{subfig:tripletloss}, the hard triplet loss  focuse on reducing the intra-class distance and increasing the inter-class distance, which is effective in the conventional retrieval task. However, the hard triplet loss does not perform very well in RGB-IR person re-identification task. As shown in Fig.\ref{subfig:outward-b}, the same  person in different modalities can be dissimilar. The hard triplet loss does not consider cross-modality factors, and hence  the training model does not deal well with cross-modality and intra-modality variations at the same time.

To address the huge cross-modality and intra-modality variations in cross-class or intra-class, we propose a hard global triplet loss based on a cross-modality batch (cm-batch) structure. Specifically, in each cm-batch, $P$ individuals are randomly selected, each person randomly selects $K$ RGB images and $K$ infrared images. For an anchor image $x^{a}_{i}$, the sum of cross-modality negative set $x^{cn}$ and intra-modality negative set $x^{in}$ constitutes the global negative set $x^{n}$, and the sum of cross-modality positive set $x^{cp}$ and intra-modality positive set $x^{ip}$ constitutes the global positive set $x^{p}$. The hard global triplet loss is computed as follows,
\begin{align}\label{HardGlobalTriplet}
    L_{hgt}=\
     \overbrace{\sum_{i=1}^{P}\sum_{a=1}^{2K}}^{all\ anchor}\
     [\alpha + \overbrace{\max_{\substack {p=1\dots 2K \\ p\neq a}}d(x^{a}_{i},x^{p}_{i})}^{hardest\ global\ positive}& \notag \\
     -\underbrace{\min_{\substack {n=1\dots 2K \\ j=1\dots P \\j\neq i }}d(x^{a}_{i},x^{n}_{j})}_{hardest\ global\ negative}&]_{+}
\end{align}
where $\alpha$ is a hyperparameter, and $x^{a}_{i} \in x^{a},x^{p}_{i} \in x^{p},x^{n}_{i} \in x^{n}$, $x^{j}_{i}$ represents the i-th image of the j-th person in the corresponding set of anchor. For any $x^{a}_{i}$ in the cm-batch, the hardest global positive or negative may be the same  or different modality.

Although hard global triplet loss can handle cross-modality and intra-modality variations at the same time, usually cross-modality variations are much larger than intra-modality variations. We thus design a hard cross-modality loss to  handle cross-modality variations. The hard cross-modality triplet loss is computed as follows,
\begin{align}\label{HardCrossTriplet}
    L_{hct}=\
     \overbrace{\sum_{i=1}^{P}\sum_{a=1}^{2K}}^{all\ anchor}\
     [\alpha + \overbrace{\max_{\substack {cp \in A}}d(x^{a}_{i},x^{cp}_{i})}^{hardest\ cross-modality\ positive}& \notag \\
     -\underbrace{\min_{\substack {cn \in A \\k=1\dots K \\k\neq i }}d(x^{a}_{i},x^{cn}_{k})}_{hardest\ cross-modality\ negative}&]_{+}
\end{align}
where ${A=\{1,2,\dots,K\}}$ when ${a\ge K}$, and otherwise \\ ${A=\{K+1,K+2,\dots,2K\}}$. Consistent with Equation \eqref{HardGlobalTriplet}, $x^{j}_{i}$ represents the i-th image of the j-th person in the corresponding set of anchor.

Our proposed hard pentaplet loss consists of hard global and cross-modality
loss. For an anchor image $x^{a}_{i}$ in cm-batch, the hardest global
triplet pair ${\{x^{a}_{i},x^{p}_{j},x^{n}_{k}\}}$ and the hardest
cross-modality triplet pair ${\{x^{a}_{i},x^{cp}_{h},x^{cn}_{t}\} }$ can be
obtained by hard sampling methods, i.e., combining the hardest triplet pairs
above to obtain a hardest pentaplet pair
${\{x^{a}_{i},x^{p}_{j},x^{n}_{k},x^{cp}_{h},x^{cn}_{t}\}}$. Note that
$x^{p}_{j}$ and $x^{cp}_{h}$ , $x^{n}_{k}$ and $x^{cn}_{t}$ may be the same
image. The hard pentaplet loss can be expressed as follows,
\begin{align}\label{HardPentaplet}
    L_{HP}=\frac{1}{2 \times P \times K}(L_{hgt} + L_{hct})
\end{align}

As shown in Fig.\ref{subfig:pentapletloss}, after the training of hard pentaplet loss, the distribution of human images in Euclidean space is more discriminative. The hard pentaplet loss has two main advantages: 1) The hard pentaplet loss can handle intra-modality  and deeper cross-modality variations simultaneously. 2) The hard pentaplet sampling method uses a limited number of images to generate sufficient hardest pentaplet pairs, which enriches the training samples and speeds up model convergence.

\subsection{Hard pentaplet with identity loss}
We use the identity loss to handle intra-class variations. As shown in Fig.\ref{subfig:outward-b}, \ref{subfig:outward-d}, there may be large variations in person images of the same identity. Given the success of identity loss in cross-modality Re-ID task, identity loss enables the CNN framework to extract the identity-specific information to reduce intra-class variations. We regard the same person in the heterogeneous modality as the same class, and the identity loss is then expressed by softmax loss, as follows,
\begin{align}\label{IdLoss}
    L_{id}=\frac{1}{2 \times P \times K}\sum_{i=1}^{2PK}-\log (\frac{e^{f_{y_{i}}}}{\sum_{j}{e^{f_{j}}}})
\end{align}
where $2PK$ is the number of training samples in cm-batch,  $f$ is designed as the output vector of the last fully connected layer in CNN, $f_{j}$ denotes the j-th element of class score vector $f$,  $j \in [1,T]$, $T$ is the number of class, $y_{i}$ is the class label of the input image $x_{i}$, and $f_{y_{i}}$ is the class score of $x_{i}$.

We add identity loss to our framework to learn a more robust feature representation. HPI loss are combined by hard pentaplet loss and identity loss, which can be expressed as follows,
\begin{align}\label{IdLoss}
    L_{HPI}=L_{hp} + L_{id}
\end{align}

\section{Experimental Results }\label{sec:Experimental}

In this section, we conduct a series of experiments to evaluate the effectiveness of the proposed method.

\subsection{Datasets and settings}

The publicly available SYSU-MM01 dataset are adopted for evaluation, which is the first benchmark for RGB-IR Re-ID. As shown in Fig.\ref{fig:SYSU}, the SYSU-MM01 dataset contains 491 identities with  287628 RGB images and 15792 infrared images in total, captured by four RGB cameras and two IR cameras. RGB cameras work in bright environments while IR cameras work in dark. Camera 1, 2, 3 capture indoor images, and camera 4, 5, 6 capture outdoor images.

\begin{table*}[!htbp]
\centering
\resizebox{0.99\textwidth}{!}{
\begin{tabular}{|c|c|c|c|c|c|c|c|c||c|c|c|c|c|c|c|c|}

\hline
\multicolumn{1}{|c|}{ \multirow{3}*{Method} }& \multicolumn{8}{c||}{All-search}&\multicolumn{8}{c|}{Indoor-search}\\ 

\cline{2-17}
\multicolumn{1}{|c|}{}& \multicolumn{4}{c|}{Single-shot}&\multicolumn{4}{c||}{Multi-shot}&\multicolumn{4}{c|}{Single-shot}&\multicolumn{4}{|c|}{Multi-shot}\\

\cline{2-17}
\multicolumn{1}{|c|}{} &  r1   &  r10  &  r20  &  mAP  &   r1  &  r10  &  r20  &  mAP  &  r1   &  r10  &  r20  &  mAP  &   r1  &  r10  &  r20  & mAP \\
\hline
HOG+Euclidean          &  2.76 & 18.25 & 31.91 &  4.24 &  3.82 & 22.77 & 37.63 &  2.16 &  3.22 & 24.68 & 44.52 &  7.25 &  4.75 & 29.06 & 49.38 & 3.51 \\
\hline
HOG+CRAFT              &  2.59 & 17.93 & 31.50 &  4.24 &  3.58 & 22.90 & 38.59 &  2.06 &  3.03 & 24.07 & 42.89 &  7.07 &  4.16 & 27.75 & 47.16 & 3.17 \\
\hline
HOG+CCA                &  2.74 & 18.91 & 32.51 &  4.28 &  3.25 & 21.82 & 36.51 &  2.04 &  4.38 & 29.96 & 50.43 &  8.70 &  4.62 & 34.22 & 56.28 & 3.87 \\
\hline
HOG+LFDA               &  2.33 & 18.58 & 33.38 &  4.35 &  3.82 & 20.48 & 35.84 &  2.20 &  2.44 & 24.13 & 45.50 &  6.87 &  3.42 & 25.27 & 45.11 & 3.19 \\
\hline
LOMO+CCA               &  2.42 & 18.22 & 32.45 &  4.19 &  2.63 & 19.68 & 34.82 &  2.15 &  4.11 & 30.60 & 52.54 &  8.83 &  4.86 & 34.40 & 57.30 & 4.47 \\
\hline
LOMO+CRAFT             &  2.34 & 18.70 & 32.93 &  4.22 &  3.03 & 21.70 & 37.05 &  2.13 &  3.89 & 27.55 & 48.16 &  8.37 &  2.45 & 20.20 & 38.15 & 2.69 \\
\hline
LOMO+CDFE              &  3.64 & 23.18 & 37.28 &  4.53 &  4.70 & 28.23 & 43.05 &  2.28 &  5.75 & 34.35 & 54.90 & 10.19 &  7.36 & 40.38 & 60.33 & 5.64 \\
\hline
LOMO+LFDA              &  2.98 & 21.11 & 35.36 &  4.81 &  3.86 & 24.01 & 40.54 &  2.61 &  4.81 & 32.16 & 52.50 &  9.56 &  6.27 & 36.29 & 58.11 & 5.15 \\
\hline
One-stream             & 12.04 & 49.68 & 66.74 & 13.67 & 16.26 & 58.14 & 75.05 &  8.59 & 16.94 & 63.55 & 82.10 & 22.95 & 22.62 & 71.74 & 87.82 & 15.04 \\
\hline
Two-stream             & 11.65 & 47.99 & 65.50 & 12.85 & 16.33 & 58.35 & 74.46 & 8.03  & 15.60 & 61.18 & 81.02 & 21.49 & 22.49 & 72.22 & 88.61 & 13.92 \\
\hline
zero-padding           & 14.80 & 54.12 & 71.33 & 15.95 & 19.13 & 61.40 & 78.41 & 10.89 & 20.58 & 68.38 & 85.79 & 26.92 & 24.43 & 75.86 & 91.32 & 18.64 \\
\hline
cmGAN                  & 26.97 & 67.51 & 80.56 & 27.80 & 31.49 & 72.74 & 85.01 & 22.27 & 31.63 & 77.23 & 89.18 & 42.19 & 37.00 & 80.94 & 92.11 & 32.76 \\
\hline
BDTR                   & 17.01 & 55.43 & 71.96 & 19.66 &   /   &   /   &   /   &   /   &    /  &   /   &   /   &   /   &   /   &   /   &   /   &   /   \\
\hline

IPVT-1+MSR             & 23.18 & 51.21 & 61.73 & 22.49 &   /   &   /   &   /   &   /   &    /  &   /   &   /   &   /   &   /   &   /   &   /   &   /   \\
\hline
D$^{2}$RL             & 28.9  & 70.6  & 82.4  & 29.2  &   /   &   /   &   /   &   /   &    /  &   /   &   /   &   /   &   /   &   /   &   /   &   /   \\

\hline
Res-Mid+\textbf{HPI}   & 40.49 & 83.61 & 93.13 &  41.64  & \textbf{47.70}& 87.99 & 95.34 &  35.15 & 45.65 & 90.76& 97.77 &  56.19 & 50.79 & 93.03 & 97.86 & 46.21 \\ 
\hline
MGN+\textbf{HPI}       & 39.77 & 79.78 & 90.14 &  41.12  &  44.86 & 82.54 & 91.61 &  34.88  &  44.06  & 87.77 & 95.59 &  54.52   &   50.55   & 89.99 & 96.06 &   44.90   \\
\hline
PCB+\textbf{HPI}       & 33.29 & 80.66 & 91.42 & 35.15  &  38.55 & 82.86 & 92.82 &  28.16  &  39.70  & 88.26 & 96.68  &  50.49   &  46.86    & 90.31 & 96.85 &   40.93     \\
\hline
MLFN+\textbf{HPI}      & 33.34 & 78.54 & 89.66 &  36.13  &  39.45 & 83.21 & 92.45 &  29.52  &  36.25  & 85.07   &  94.51  &  47.99   &  41.99    & 86.34 & 95.20 &   38.43     \\
\hline
BFE+\textbf{HPI}      & \textbf{41.36} &  \textbf{84.78} & \textbf{94.51}  &  \textbf{42.95}  &  47.56 &  \textbf{88.13}  &  \textbf{95.98}  &  \textbf{36.08}  &  \textbf{45.77}  & \textbf{91.82}  &  \textbf{98.46}  &  \textbf{56.52}   &  \textbf{53.05}    &  \textbf{93.71} & \textbf{98.93} &   \textbf{47.48}     \\
\hline
\end{tabular}
}
\caption{Comparison with the state-of-the-arts on the SYSU-MM01 dataset.
r1, r10, r20 denote rank-1, 10, 20 accuracies(\%).} \label{tab:results}

\end{table*}

\subsection{Evaluation protocol}
The SYSU-MM01 dataset is divided into training set and test set, where the former  contains 395 persons with 22258 RGB images and 11909 infrared images, and the latter  contains 96 persons. Note that a person does not appear in the two sets simultaneously.

In the training stage, all images in the training set can be used for training. In the test stage, the RGB images are  for the gallery set and the infrared images are  for the probe set. There are two verification modes: all-search mode and indoor-search mode. For the all-search mode, the RGB images from RGB cameras 1, 2, 4 and 5 are for the gallery set and the infrared images from IR cameras 3 and 6 are for the probe set. For the indoor-search mode, the RGB images from RGB cameras 1 and 2 are for the gallery set and the infrared images from IR cameras 3 are for the probe set. For each mode, there are multi-shot and single-shot settings. For every identity in gallery set, we randomly select 1/10 images from the RGB camera as single-shot/multi-shot setting respectively. For the probe set, all infrared images are used.

For a given probe image, we match it by calculating the similarity between the probe image and gallery images. The matching of the Re-ID is performed between cameras at different positions, so the probe images of  camera 3 skips the gallery images of  camera 2 because camera 2 and  camera 3 are located at the same position. After calculating the similarity, we can obtain the ranking list according to the descending order of similarity. To indicate the performance, we use Cumulative Match Characteristic curve (CMC)\cite{moon2001computational} and average accuracy (mAP).

\subsection{Implementation details}

We use NVIDIA GeForce 1080Ti graphics cards with Pytorch computing framework to implement our algorithm. Five RGB-RGB Re-ID neural networks were used to verify the superiority of our algorithms: Res-Mid, MGN, PCB, BFE, MLFN, which are described in Section \ref{sec:ReIDNetwork}. As shown in Table \ref{tab:modelset}, the input image size and the output embedding feature dimension are different due to the difference of the model. The infrared image is padding to three channels, which copies the information of one channel. We use the Adam \cite{kingma2014adam} optimizer to train 10$k$ iterations, and the initial learning rate is set to $3*10^{-4}$.

\begin{table}[]
\centering
\resizebox{0.49\textwidth}{!}{
\begin{tabular}{c|c|c|c|c}
\hline
Model   & W*H     & Dim   & Batch Size & Lr          \\ \hline
Res-Mid & 224*224 & 3072  & 64         & $3*10^{-4}$ \\
MGN     & 128*384 & 2048  & 64         & $3*10^{-4}$ \\
PCB     & 224*224 & 12288 & 64         & $3*10^{-4}$ \\
BFE     & 128*256 & 1024  & 64         & $3*10^{-4}$ \\
MLFN    & 224*224 & 1024  & 64         & $3*10^{-4}$ \\ \hline
\end{tabular}
}
\caption{Settings for different model training: input image width and height (W*H), feature dimension (Dim), training batch size (Batch Size) and learning rate (Lr).}
\label{tab:modelset}
\end{table}

Since our hard pentaplet loss requires slightly different cm-batches, we sample a $2PK$ batch by randomly sampling $P$ identities, and each person randomly samples $K$ RGB images and $K$ infrared images. In our experiment, $P$ is set to 8, $K$ is set to 4, and the batch size is calculated to be 64. For input images, the methods of random horizontal flip and random cropping is used to expand the amount of data. We set margin $\alpha$ in hard pentaplet loss in the range [0.3, 0.6, 0.9, 1.2, 1.5, 1.8] and evaluate our method by experimenting with other hyper-parameters.

\begin{table}[]
\centering
\resizebox{0.49\textwidth}{!}{
\begin{tabular}{c|c|c|c|c|c|c}
\hline
        & \multicolumn{2}{c|}{Market1501} & \multicolumn{2}{c|}{CUHK03} & \multicolumn{2}{c}{DukeMTMC-reID} \\ \cline{2-7}
Method  & r1             & mAP            & r1           & mAP          & r1            & mAP           \\ \hline   \hline
Res-Mid & 89.87          & 75.55          & 43.51        & 47.14        & 63.88         & 80.43         \\
MGN     & 95.7           & 86.9           & 66.8         & 66           & 88.7          & 78.4          \\
PCB     & 92.4           & 77.3           & 61.3         & 54.2         & 81.9          & 65.3          \\
BFE     & 94.4           & 85             & 72.1         & 67.9         & 88.7          & 75.8          \\
MLFN    & 90             & 74.3           & 52.8         & 47.8         & 81.0          & 62.8          \\ \hline
\end{tabular}
}
\caption{Performance of RGB-RGB Re-ID models on Market1501, CUHK03, DukeMTMC-reID datasets.}
\label{tab:RGB-RGBmodel}
\end{table}

\subsection{Comparison with the state-of-the-arts}
We evaluated our HPILN method against 15 existing methods on the SYSU-MM01 dataset in Table \ref{tab:results}. For performance measure, the rank-1, 10, 20 accuracies of Cumulative Match Characteristic curve (CMC) and mean average precision (mAP) are used to show the clear performance superiority of our method. The comparison contains four state-of-the-art methods:

\begin{itemize}
\item \textbf{Zero-padding}\cite{wu2017rgb}. A deep zero-padding method
    for training one-stream network towards automatically capturing
    domain-specific information for cross-modality matching.

\item \textbf{BDTR}\cite{ye2018visible}. A dual-path network with
    bi-directional dual-constrained top-ranking loss to learn
    discriminative feature representations from two modalities.

\item \textbf{CmGAN}\cite{dai2018cross}. A cross-modality generative
    adversarial network  using a cutting-edge generative adversarial
    training based discriminator and cross-modality triplet loss to learn
    discriminative feature representation from two modalities.

\item \textbf{IPVT-1+MSR}\cite{kang2019person}. IPVT-1 combining RGB image
and infrared as a single input to reduce computational complexity.
Moreover, the accuracy of Re-ID is improved by multi-scale Retinex (MSR)-filtered input images.

\item \textbf{D$^{2}$RL}\cite{Wang_2019_CVPR}. Dual-level discrepancy reduction learning (D$^{2}$RL) to
decompose and handle the mixed  modality and appearance discrepancies. Images from different modalities
 are mapped to a unified space, and then a cascaded sub-network is used to obtain discriminative features.
\end{itemize}

In addition, other existing methods are used for comparison, including handcrafted features such as  HOG\cite{dalal2005histograms} and LOMO\cite{liao2015person}, cross-domain models such as CDFE\cite{lin2006inter} and CRAFT\cite{chen2018person}, CCA\cite{rasiwasia2010new}, one-stream and two-stream networks\cite{wu2017rgb}, and metric learning method LFDA\cite{pedagadi2013local}. Most of the results were obtained from the references \cite{wu2017rgb,dai2018cross,ye2018visible,kang2019person,Wang_2019_CVPR}.

We use Res-Mid\cite{yu2017devil}, MGN\cite{wang2018learning}, PCB\cite{sun2018beyond}, BFE\cite{dai2018batch}, MLFN\cite{chang2018multi} as feature extractors in our HPILN method. To our best knowledge, these models are the state-of-the-art methods in RGB-RGB Re-ID in the past two years, and Table \ref{tab:RGB-RGBmodel} shows their performance on the Market1501\cite{zheng2015scalable}, CUHK03\cite{li2014deepreid} and DukeMTMC-reID\cite{ristani2016performance} datasets.

In Table \ref{tab:results}, the results of five rows on the bottom show the performance of HPILN method which applies HPI loss to five models. It is clear that our HPILN method is significantly better than all existing methods in the SYSU-MM01 benchmark, where the five models based on HPI loss have higher rank-1, 10, 20 and mAP in all verification modes and setting than existing methods. Specifically, the BFE model based HPI loss performs the best in most of the indicators, which outperforms the 2nd best method (D$^{2}$RL) on all-search single-shot setting in terms of the rank1 and mAP metric 12.46\% (41.36-28.9) and 13.75\% (42.95-29.2), respectively.

\begin{figure}[!b]
\centering{
\subfigure[Identity Loss]{
\begin{minipage}[t]{0.3\linewidth}
\centering
\includegraphics[width=1\linewidth]{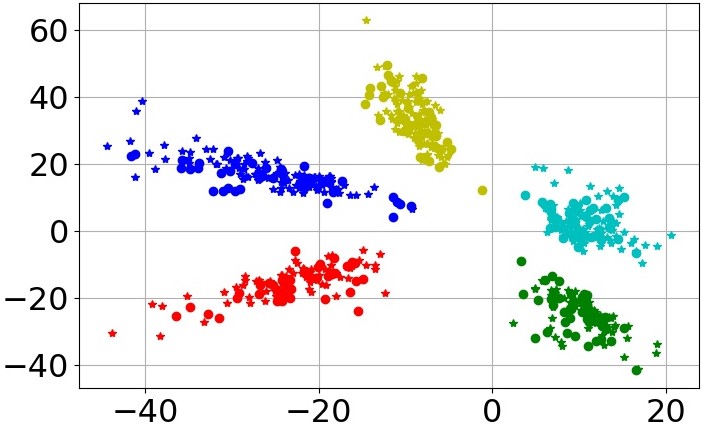}
\label{subfig:identityloss}
\end{minipage}%
}
\subfigure[HP Loss]{
\begin{minipage}[t]{0.3\linewidth}
\centering
\includegraphics[width=1\linewidth]{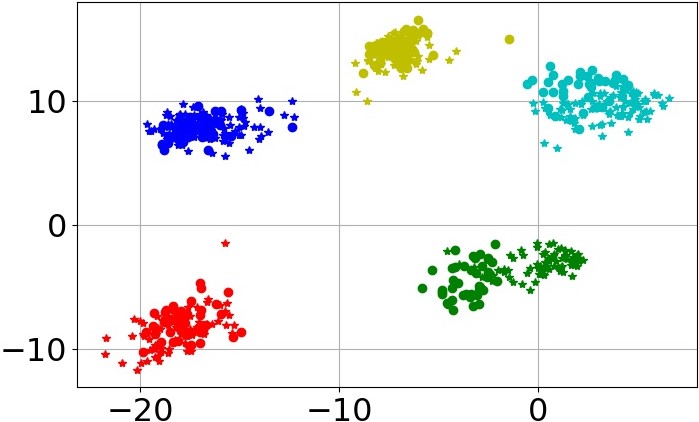}
\label{subfig:HPloss}
\end{minipage}%
}%
\subfigure[HPI Loss]{
\begin{minipage}[t]{0.3\linewidth}
\centering
\includegraphics[width=1\linewidth]{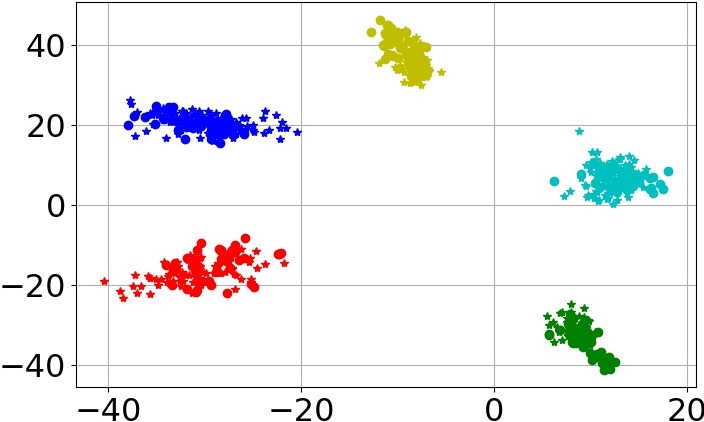}
\label{subfig:HPIloss}
\end{minipage}%
}}
\centering \caption{Comparison among identity loss, HP loss and HPI loss. In this toy experiment, we modified Res-Mid to learn a 2-D feature on a subset of the SYSU-MM01 dataset. Specifically, we set the output of dimension of the last fully connected layer as 2 and visualize the learned features. The five color points represent five identity classes, the circular and star shapes represent RGB modality and IR modality respectively.
\label{fig:lossfeature}}
\end{figure}

\subsection{Effectiveness of fusion loss}

To verify the effectiveness of fusion identity loss and hard pentaplet loss, we compared the rank-1 precision of identity loss, hard pentaplet (HP) loss and hard pentaplet with identity (HPI) loss on the SYSU-MM01 dataset. We report the results with five models in Table \ref{tab:idlosscompare}. As can be seen from Table \ref{tab:idlosscompare}, the combination of identity loss is effective. It is clear that the RGB-RGB Re-ID models based on identity loss can also achieve excellent precision, even the Res-Mid based identity loss performance is better than the 2nd best method (cmGAN) in rank-1 accuracies. In addition, although HP loss has shown excellent performance, HPI loss which integrates identity loss and HP loss further improves the accuracy. We speculate that the fusion of identity loss further enhances the feature discrimination of HP loss.

In order to verify the above speculation, we conducted a toy experiment to illustrate the differences of features in 2-D Euclidean space learned by identity loss, HP loss, and HPI loss respectively, shown in Fig.\ref{fig:lossfeature}. Under the supervision of identity loss, the learned features are slightly separable which are not discriminative enough, since Fig.\ref{subfig:identityloss} still shows large cross-modality variations and small inter-class discrimination. Fig.\ref{subfig:HPloss} shows that there is a large margin between the dot clusters, which means HP loss learned discriminative large-margin features. For HPI loss combined with HP loss and identity loss, Fig.\ref{subfig:HPIloss} shows that the same classes are clustered together and there is significant separation between the different classes. The reason why the HPI loss performance superior is that HP loss handle the cross-modality and intra-modality variations to learn the distinguishing large-margin features, and identity loss assists HP loss to further reduce intra-class distance.
\begin{table}[]
\resizebox{0.49\textwidth}{!}{
\begin{tabular}{lcccc}
\hline
\multicolumn{1}{l|}{Mode}    & \multicolumn{2}{c|}{All-search}                                    & \multicolumn{2}{c}{Indoor-search}            \\ \hline
\multicolumn{1}{l|}{Setting} & \multicolumn{1}{c|}{Single-shot} & \multicolumn{1}{c|}{Multi-shot} & \multicolumn{1}{c|}{Single-shot} & Multi-shot \\ \hline   \hline
\multicolumn{5}{l}{Res-Mid}                                                                                                                       \\ \hline
Identity Loss                & 32.68                            & 38.58                           & 37.41                            & 45.23      \\
HP Loss                      & 36.06                            & 42.32                           & 40.46                            & 44.08      \\
HPI Loss                     & 40.49                            & 47.70                           & 45.65                            & 50.79      \\ \hline
\multicolumn{5}{l}{MGN}                                                                                                                           \\ \hline
Identity Loss                & 27.29                            & 31.05                           & 33.47                            & 38.19      \\
HP Loss                      & 36.68                            & 41.62                           & 41.95                            & 48.37      \\
HPI Loss                     & 39.77                            & 44.86                           & 44.06                            & 50.55      \\ \hline
\multicolumn{5}{l}{PCB}                                                                                                                           \\ \hline
Identity Loss                & 11.22                            & 15.67                           & 8.7                              & 12.76      \\
HP Loss                      & 26.51                            & 32.40                           & 33.61                            & 40.42      \\
HPI Loss                     & 33.29                            & 38.55                           & 39.70                            & 46.86      \\ \hline
\multicolumn{5}{l}{MLFN}                                                                                                                          \\ \hline
Identity Loss                & 28.44                            & 33.23                           & 30.19                            & 34.82      \\
HP Loss                      & 30.62                            & 35.43                           & 31.28                            & 36.69      \\
HPI Loss                     & 33.34                            & 39.45                           & 36.25                            & 41.99      \\ \hline
\multicolumn{5}{l}{BFE}                                                                                                                           \\
Identity Loss                & 25.69                            & 32.02                           & 29.65                            & 36.68      \\
HP Loss                      & 38.89                            & 45.69                           & 44.51                            & 52.51      \\
HPI Loss                     & 41.36                            & 47.56                           & 45.77                            & 53.05      \\ \hline
\end{tabular}
}
\caption{Effectiveness of fusion loss on the SYSU-MM01 dataset. Rank-1 accuracies (\%) in all/indoor-search mode and single/multi-shot setting.}
\label{tab:idlosscompare}
\end{table}

\begin{figure}[!t]
\centering{
\subfigure[Rank-1]{
\begin{minipage}[t]{0.7\linewidth}
\centering
\includegraphics[width=1\linewidth]{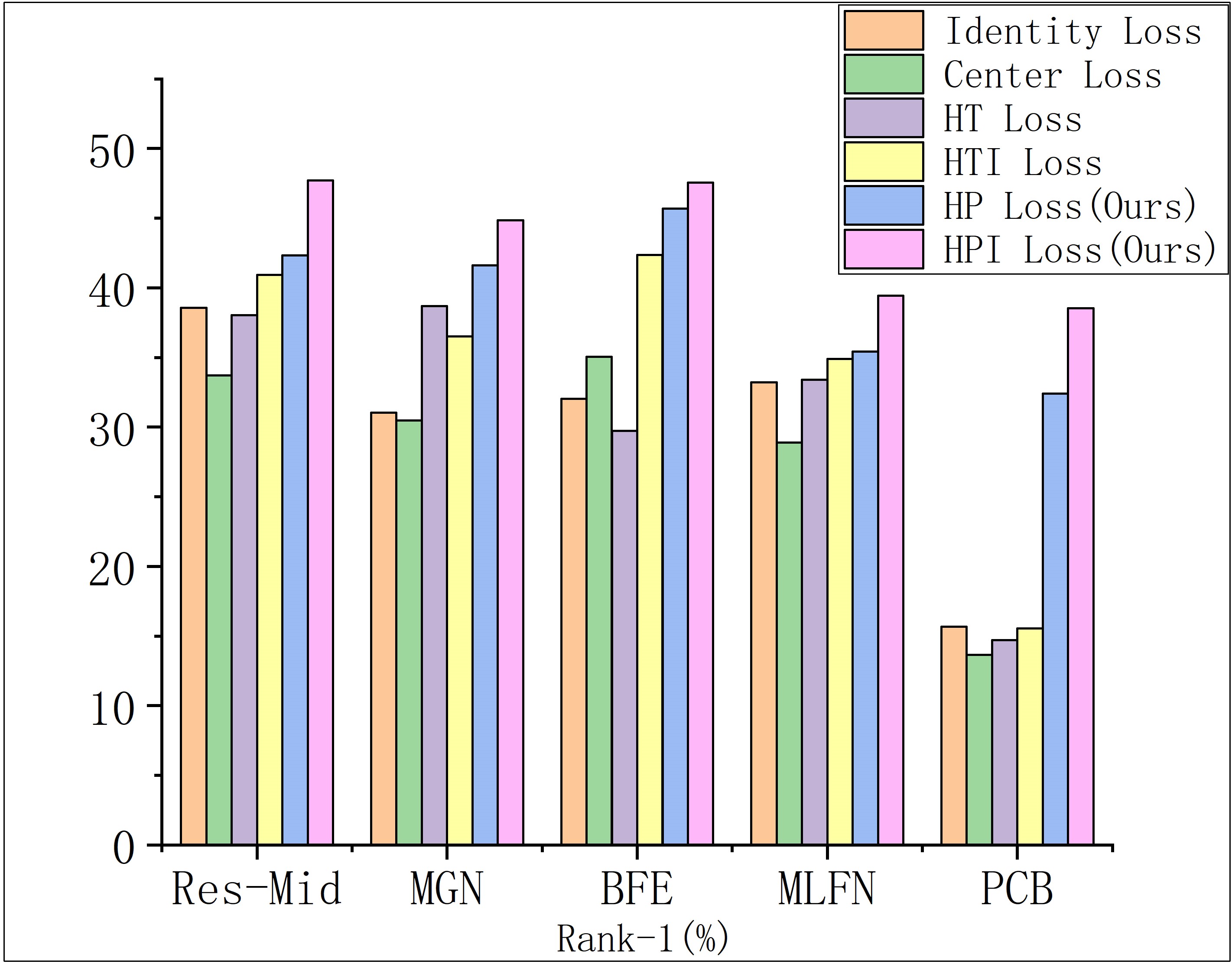}
\end{minipage}%
}%

\subfigure[mAP]{
\begin{minipage}[t]{0.7\linewidth}
\centering
\includegraphics[width=1\linewidth]{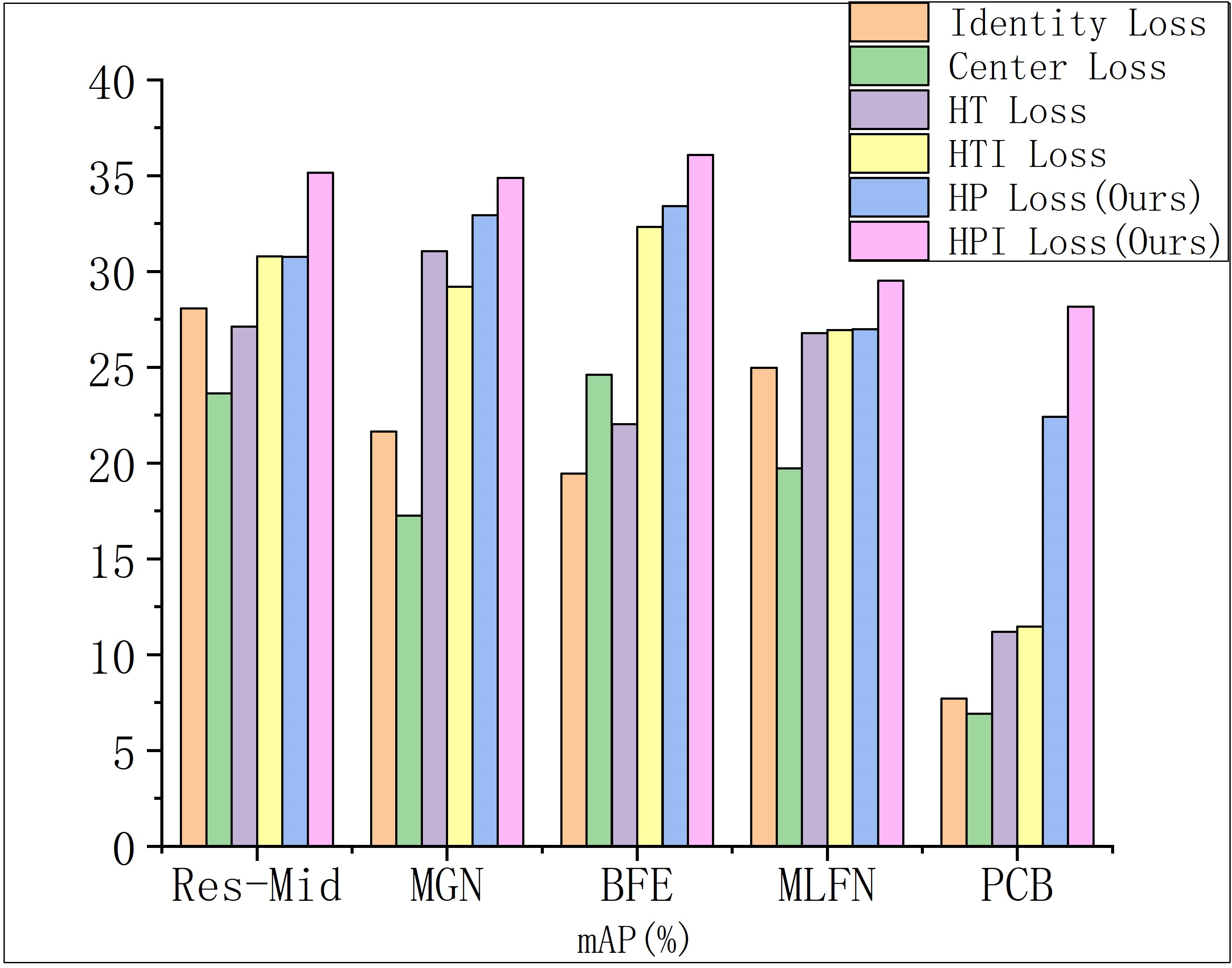}
\end{minipage}%
}}

\caption{Performance of different loss functions. We tested rank-1 and mAP with all-search multi-shot setting on five models.
\label{fig:losscompare}}

\end{figure}

\begin{figure*}[!t]
\centering{
\subfigure[Res-Mid]{
\begin{minipage}[t]{0.33\linewidth}
\centering
\includegraphics[width=1\linewidth]{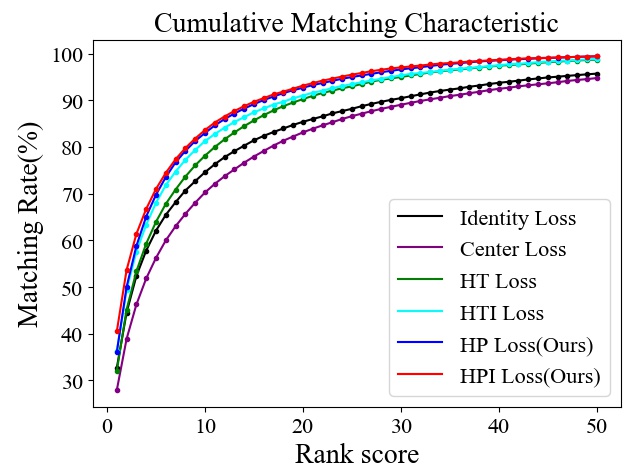}
\end{minipage}%
}%
\subfigure[MGN]{
\begin{minipage}[t]{0.33\linewidth}
\centering
\includegraphics[width=1\linewidth]{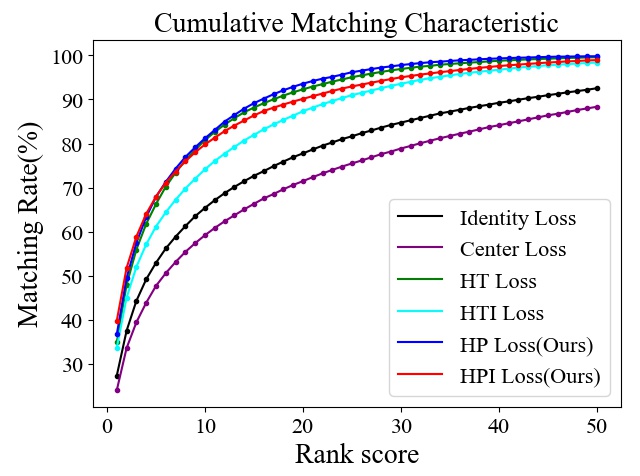}
\end{minipage}%
}%
\subfigure[BFE]{
\begin{minipage}[t]{0.33\linewidth}
\centering
\includegraphics[width=1\linewidth]{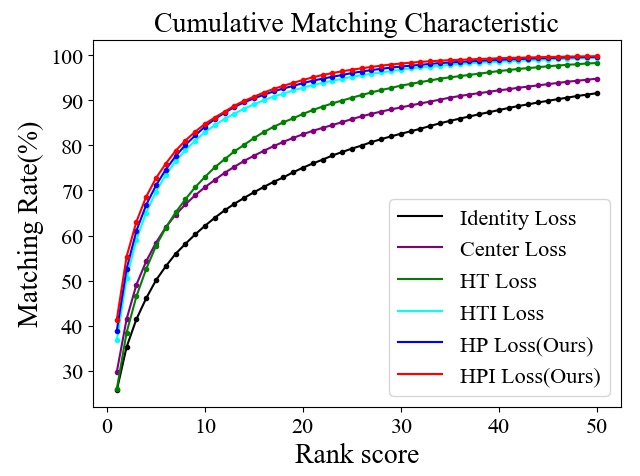}
\end{minipage}%
}%

\subfigure[MLFN]{
\begin{minipage}[t]{0.33\linewidth}
\centering
\includegraphics[width=1\linewidth]{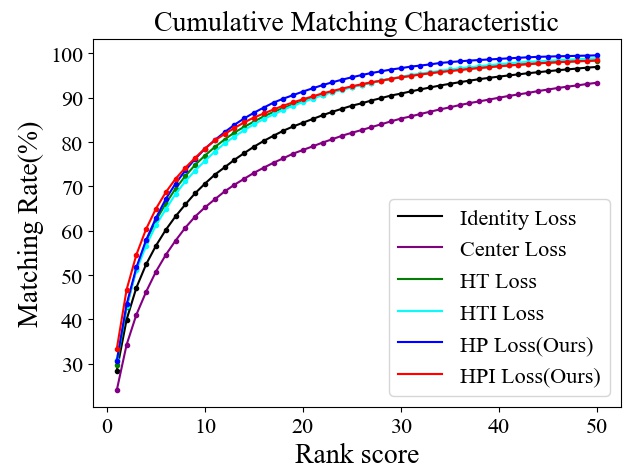}
\end{minipage}%
}%
\subfigure[PCB]{
\begin{minipage}[t]{0.33\linewidth}
\centering
\includegraphics[width=1\linewidth]{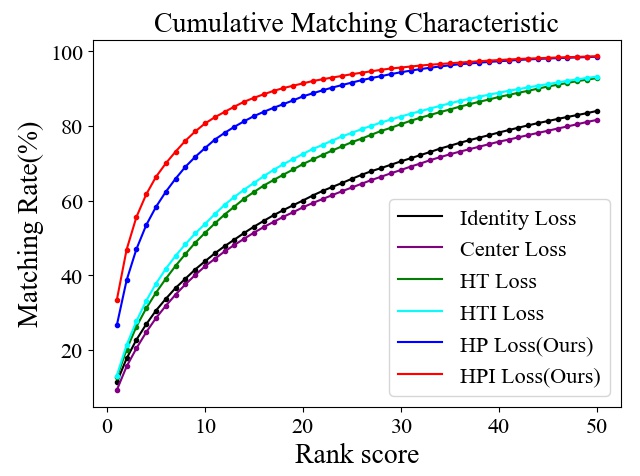}
\end{minipage}%
}}
\caption{ Cumulative Match Characteristic curve of different models under different loss functions \label{fig:CMC}}
\end{figure*}

\subsection{Comparison with other advanced loss}

To demonstrate the superiority of our methods, we compare our HP loss and HPI loss with other advanced loss functions in Re-ID, including HT (hard triplet) loss \cite{hermans2017defense}, HTI (hard triplet with identity) loss, center loss \cite{wen2016discriminative}, and identity loss \cite{xiao2016learning}. The performance of the contrast methods was reported in Fig.\ref{fig:losscompare} and Fig.\ref{fig:CMC}.

Results shown in Fig.\ref{fig:losscompare} illustrate that HP loss and HPI loss have better performance on rank-1 and mAP than other loss functions. We tested five models, and the rank-1 and mAP of the PCB based HPI loss were 22.88\% and 16.69\% higher than the second loss (except HP loss), respectively.

Fig.\ref{fig:CMC} shows the Cumulative Match Characteristic (CMC) curves of different models under different loss in the SYSU-MM01 dataset. The CMC curve can more fully reflect the performance of the model. We tested five models under the all-search single-shot setting. In all tested models, HP loss and HPI loss performed better than other losses, and our method is not only higher than the existing method in rank-1, but also maintains a lead in rank 1-50.

The reason we are better than other methods is that we consider cross-modality variations to better extract common features in heterogeneous modality. In addition, we use a more reasonable image sampling method to balance the number of input images so that the model does not focus on a certain modality images.

\subsection{Analysis of model selection}
In HPILN framework, the RGB-RGB Re-ID models was first adopted as the feature extractor for RGB-IR Re-ID task. We tested the performance of the classification models and some RGB-RGB Re-ID models in HPILN framework. The RGB-RGB Re-ID models include different types: global-based networks, part-based networks and attention-based networks. The results are shown in Table.\ref{tab:othermodel}.

In the HPILN framework, the RGB-RGB Re-ID model is more suitable for RGB-IR Re-ID tasks than classification models. We chose Resnet50 \cite{he2016deep} and Densenet121 \cite{huang2017densely} as classification models, which perform well on ImageNet dataset. From Table \ref{tab:othermodel}, we observed that classification models do not achieve good accuracy in the HPILN framework compared to the RGB-RGB Re-ID model. The reason is that the RGB-RGB Re-ID model is designed for person images. Although infrared images and RGB images are very different, heterogeneous images also have certain common features, such as body shape and clothing shape. Therefore, the RGB-RGB Re-ID model performs well in RGB-IR Re-ID tasks.

However, not all RGB-RGB Re-ID models perform well in RGB-IR Re-ID tasks. We tested two attention-based RGB-RGB Re-ID models: mudeep \cite{qian2017multi} and hacnn \cite{li2018harmonious}. From Table \ref{tab:othermodel}, mudeep and hacnn have lower precision on SYSU-MM01. Both mudeep and hacnn use the attention mechanism which automatically focus on local salient areas for computing deep features. Attention mechanism is not robust in cross-modality training because there are few similar local regions of heterogeneous images.

\begin{table}[]
\resizebox{0.49\textwidth}{!}{
\begin{tabular}{cc|cc}
\hline
                                      &             & \multicolumn{2}{c}{All-search}         \\
Type                                  & Method      & Single-shot                & Multi-shot \\ \hline   \hline
\multirow{2}{*}{Global-based}         & Res-Mid     & \multicolumn{1}{c|}{40.49} & 47.70      \\
                                      & MLFN        & \multicolumn{1}{c|}{33.34} & 39.45      \\     \hline
\multirow{3}{*}{Part-based}           & PCB         & \multicolumn{1}{c|}{33.29} & 38.55      \\
                                      & MGN         & \multicolumn{1}{c|}{39.77} & 44.86      \\
                                      & BFE         & \multicolumn{1}{c|}{41.36} & 47.56      \\    \hline
\multirow{2}{*}{Attention-based}      & Hacnn       & \multicolumn{1}{c|}{1.07}  & 1.53       \\
                                      & Mudeep      & \multicolumn{1}{c|}{9.35}  & 11.78      \\    \hline
\multirow{2}{*}{Classification-based} & Resnet50    & \multicolumn{1}{c|}{5.36}  & 5.69       \\
                                      & Densenet121 & \multicolumn{1}{c|}{13.59} & 16.10      \\ \hline
\end{tabular}
}
\centering \caption{ The rank-1 of classification models and RGB-RGB Re-ID
models in HPILN framework.} \label{tab:othermodel}
\end{table}

\section{Conclusion}\label{sec:Conclusion}

A novel feature learning framework based on hard pentaplet and identity loss network (HPILN) is proposed for RGB-IR person re-identification. In the framework, existing RGB-RGB Re-ID model is used as the feature extractor, hard pentaplet (HP) loss is used to learn the discriminative large-margin features in order to  handle cross-modality and intra-modality variations, and the identity loss is combined to extract identity-specific information to learn the separation features.  The experimental results show that our method achieves state-of-the-art performance on SYSU-MM01 dataset.

\bibliographystyle{IEEEtran}
\bibliography{thebib}

\end{document}